\documentclass[a4paper,twoside]{article}

\usepackage{epsfig}
\usepackage{subcaption}
\usepackage{calc}
\usepackage{amssymb}
\usepackage{amstext}
\usepackage{amsmath}
\usepackage{amsthm}
\usepackage{multicol}
\usepackage{pslatex}
\usepackage{apalike}
\usepackage{algorithm2e}
\usepackage[bottom]{footmisc}
\usepackage{SCITEPRESS} 
\usepackage{xcolor}
\usepackage[hidelinks]{hyperref}
\usepackage{titlesec}

\begin{document}

\title{BEVSeg2TP: Surround View Camera Bird’s-Eye-View Based Joint Vehicle Segmentation and Ego Vehicle Trajectory Prediction}

\author{\authorname{Sushil~Sharma\sup{1,2}, Arindam~Das\sup{1}, Ganesh~Sistu\sup{1}, Mark~Halton\sup{1}, and Ciarán~Eising\sup{1,2}} \vspace{0.3cm}
\affiliation{\sup{1}Department of Electronic \& Computer Engineering, University of Limerick, Ireland}
\affiliation{\sup{2}SFI CRT Foundations in Data Science, University of Limerick, Ireland}
\email{\sup{1,2}firstname.lastname@ul.ie} 
}

\ifx01 
\newcommand{\AD}[1]{\textcolor{blue}{[Arindam: #1]}}
\else 
\newcommand{\AD}[1]{\textcolor{blue}{}}
\fi

\ifx01 
\newcommand{\SKS}[1]{\textcolor{red}{[Sushil: #1]}}
\else 
\newcommand{\SKS}[1]{\textcolor{red}{}}
\fi


\abstract{Trajectory prediction is, naturally, a key task for vehicle autonomy. While the number of traffic rules is limited, the combinations and uncertainties associated with each agent's behaviour in real-world scenarios are nearly impossible to encode. Consequently, there is a growing interest in learning-based trajectory prediction. The proposed method in this paper predicts trajectories by considering perception and trajectory prediction as a unified system. In considering them as unified tasks, we show that there is the potential to improve the performance of perception. To achieve these goals, we present BEVSeg2TP - a surround-view camera bird's-eye-view-based joint vehicle segmentation and ego vehicle trajectory prediction system for autonomous vehicles. The proposed system uses a network trained on multiple camera views. The images are transformed using several deep learning techniques to perform semantic segmentation of objects, including other vehicles, in the scene. The segmentation outputs are fused across the camera views to obtain a comprehensive representation of the surrounding vehicles from the bird's-eye-view perspective. The system further predicts the future trajectory of the ego vehicle using a spatiotemporal probabilistic network (STPN) to optimize trajectory prediction. This network leverages information from encoder-decoder transformers and joint vehicle segmentation. 
The predicted trajectories are projected back to the ego vehicle’s bird’s-eye-view perspective to provide a holistic understanding of the surrounding traffic dynamics, thus achieving safe and effective driving for vehicle autonomy. The present study suggests that transformer-based models that use cross-attention information can improve the accuracy of trajectory prediction for autonomous driving perception systems. Our proposed method outperforms existing state-of-the-art approaches on the publicly available nuScenes dataset. 
This link is to be followed for the source code: \textcolor{purple}{\href{https://github.com/sharmasushil/BEVSeg2TP/}{https://github.com/sharmasushil/BEVSeg2TP/}}.
}

\onecolumn \maketitle \normalsize \setcounter{footnote}{0} \vfill

\section{INTRODUCTION}
\label{sec:introduction}
Accurate trajectory prediction is a critical capability for autonomous driving systems, playing a pivotal role in enhancing safety, efficiency, and driving policies. This technology is increasingly vital as autonomous vehicles become more prevalent on public roads, as it enables these vehicles to anticipate the movements of various road users, including pedestrians, cyclists, and other vehicles. By doing so, autonomous vehicles can proactively plan and execute safe manoeuvres, reducing the risk of potential collisions \cite{li2021intelligent,cheng2019research} and effectively navigating through complex traffic scenarios. Moreover, trajectory prediction empowers autonomous vehicles to optimise their driving behaviour, enabling smoother lane changes \cite{chen2020autonomous} 
and seamless merging to improve overall traffic flow and reduce congestion \cite{wei2021traffic}. 
Furthermore, trajectory prediction also plays a crucial role in facilitating effective communication and interaction between autonomous vehicles, human drivers, and pedestrians. By behaving predictably, autonomous vehicles can earn the trust of other road users \cite{liu2021autonomous,yang2021survey} and support other extended applications in the ADAS perception stack, such as pedestrian detection \cite{10222711,dasgupta2022spatio}, and pose estimation \cite{das2022deep}. 

\begin{figure*}[ht]
    \centering
    \includegraphics[width=0.98\textwidth]{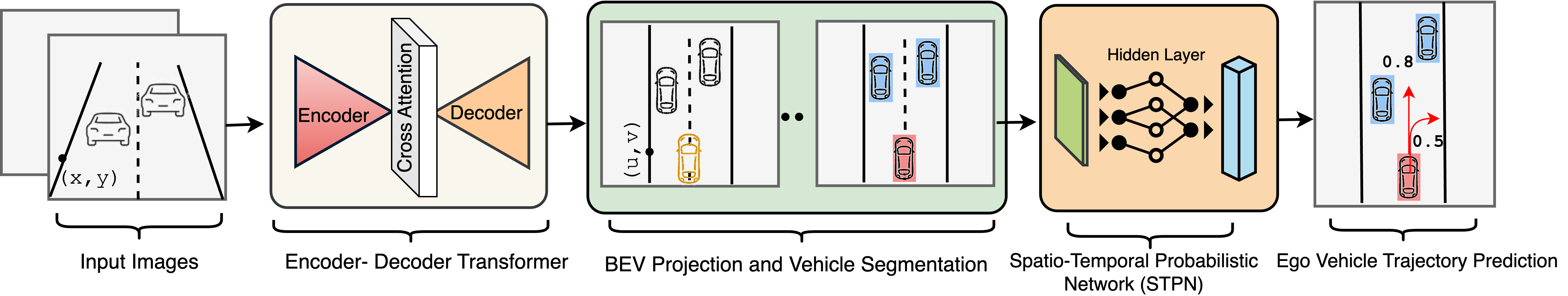}
    \caption{Our proposed \textbf{BEVSeg2TP framework - surround-view camera joint vehicle segmentation and ego vehicle trajectory prediction in bird's-eye-view} approach consists of an encoder-decoder transformer, BEV projection module followed by segmentation outputs fed to the spatio-temporal probabilistic network to produce ego vehicle trajectory prediction.}
    \label{fig:sv-sys}
\end{figure*}

In this paper, we introduce an approach called BEVSeg2TP for joint vehicle segmentation and ego vehicle trajectory prediction, leveraging a bird's-eye-view perspective from surround-view cameras. Our proposed system employs a network trained on surround-view or multi-camera view from the host vehicle, which it transforms into bird's-eye-view imagery of the surrounding context. These images undergo deep learning-driven processes to perform semantic segmentation on objects, including neighboring vehicles within the scene. The segmentation outcomes are then amalgamated across camera perspectives to generate a comprehensive representation of the surrounding vehicles from a bird's-eye-view perspective \cite{zhou2022cross}.  Building upon this segmented data, the proposed system also anticipates the future trajectories of the host vehicle using a spatio-temporal probabilistic network (STPN) \cite{cui2019multimodal}. The STPN learns the spatiotemporal patterns of vehicle motion from historical trajectory data. The predicted trajectories are then projected back to the ego vehicle's bird's-eye-view perspective to provide a holistic understanding of the surrounding traffic dynamics. 
Figure \ref{fig:sv-sys} represents the overarching depiction of our approach. Our principal contributions to the BEVSeg2TP proposal are:
\begin{itemize}
    
    \item  Our proposed deep architecture offers an approach to jointly accomplish vehicle segmentation and ego vehicle trajectory prediction tasks by combining and adapting the works of \cite{zhou2022cross,phan2020covernet,cui2019multimodal}.
    \item We propose enhancements to the capabilities of the current encoder-decoder transformer used in the spatio-temporal probabilistic network (STPN) for optimizing trajectory prediction.
    \item We implemented an end-to-end trainable surround-view camera bird's-eye-view-based network that achieves state-of-the-art results on the nuScenes dataset \cite{caesar2020nuscenes} when jointly trained with segmentation.
\end{itemize}

\section{PRIOR ART}

Joint vehicle segmentation and ego vehicle trajectory prediction using a surround or multi-camera bird's-eye view is currently an emerging area of research with several motivating factors. Firstly, working on this problem could help advance the field and contribute to the development of more effective and accurate autonomous driving systems. The potential uses of precise vehicle segmentation and predictions for ego vehicle trajectories are vast, encompassing domains such as self-driving vehicles, intelligent transportation systems, and automated driving systems, among others.

Moreover, this problem is complex and challenging, requiring the integration of information from multiple sensors and camera views. Addressing the technical challenges of this problem, such as designing effective deep learning models or developing efficient algorithms, could be a motivating factor for researchers interested in solving complex and challenging problems. Our primary focus is on enhancing map-view segmentation. It is undeniable that extensive research has been conducted in this field, which lies at the convergence of 3D recognition \cite{ma2019accurate,lai2023spherical,manhardt2019roi}, depth estimation \cite{eigen2014depth,godard2019digging,ranftl2020towards,zhou2017unsupervised}, and mapping \cite{garnett20193d,senguptaautomatic,zhu2021monocular}.

These are the key areas that can facilitate segmentation construction and improvement. While trajectory prediction or motion planning for autonomous systems is crucial, we acknowledge the need to consider various aspects of the vehicle state, such as current position and velocity, road geometry \cite{lee2016road,wiest2020learning,wu2017road}, other vehicles, environmental factors, and driver behaviour \cite{zhang2020multi,abbink2017driver,mcdonald2020drivers}. The architecture previously described by the authors \cite{sharma2023navigating} explores the utilization of the CNN-LSTM model for predicting trajectories, covering unique scenarios like pedestrians crossing roads. While the model adeptly comprehends these scenarios, it adheres to a model-driven methodology, thereby carrying inherent limitations. In our pursuit to address these limitations and devise an alternative approach, we propose the integration of a transformer-based model into our trajectory prediction methodology. Our strategy entails a partial adoption of the principles from CoverNet \cite{phan2020covernet}, albeit with notable distinctions. CoverNet's trajectory prediction relies on raster maps, whereas our model pivots towards real-time map view representations.

\section{PROPOSED METHODOLOGY}

\begin{figure*}[ht]
    \centering
    \includegraphics[width=0.95\textwidth]{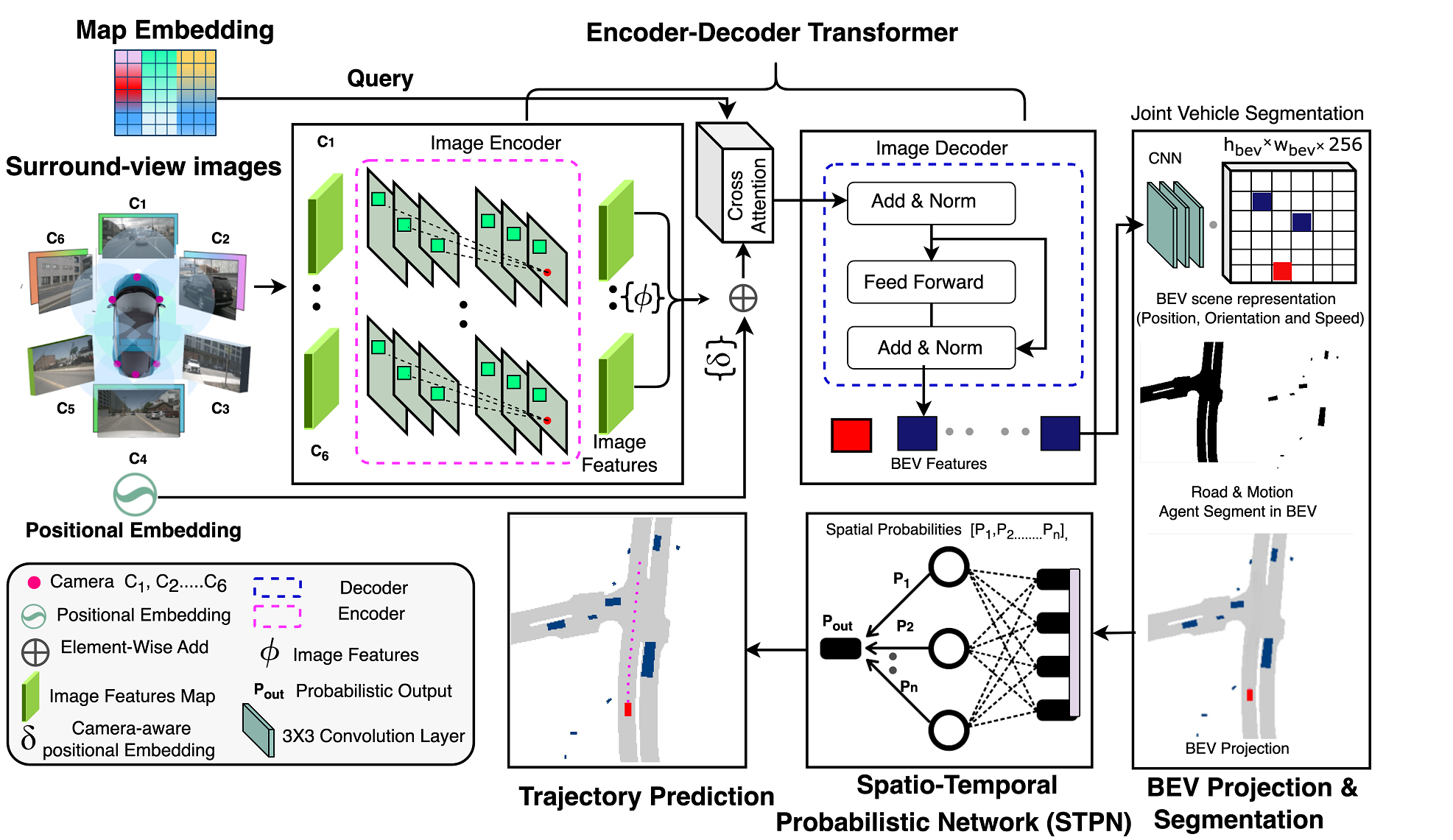}
    \caption{\textbf{Our proposed BEVSeg2TP architecture:}  Joint vehicle segmentation and ego vehicles trajectory prediction involves extracting image features $\{ \phi\}$ at multiple scales and using a camera-aware positional embedding $\{ \delta\}$ to account for perspective distortion. We then use map-view positional embedding and cross-attention layers to capture contextual information from multiple views and refine the vehicle segmentation. This segmentation information is then used as input to a spatio-temporal probabilistic network (STPN) for trajectory prediction based on the surrounding environment.}
    \label{fig:image2}
    
\end{figure*}

In this section, we present BEVSeg2TP - our proposed deep architecture designed to efficiently achieve both vehicle segmentation and ego vehicle trajectory prediction tasks simultaneously. The proposed method, as depicted in Figure \ref{fig:image2}, utilizes multiple cameras to create a comprehensive view of the environment around the ego vehicle, improving ego vehicle and object segmentation, based on the work presented by \cite{zhou2022cross}. We extend this transformer technique to incorporate trajectory prediction using a spatio-temporal probabilistic network to calculate path likelihoods, as presented in \cite{phan2020covernet,cui2019multimodal}. This approach combines multiple sources of information for more accurate future trajectory predictions, enhancing self-driving car safety and performance by jointly learning the segmentation and the trajectory prediction.

\subsection{Surround-view Camera Inputs}
The dataset used in this paper is nuScenes \cite{caesar2020nuscenes}. It consists of six cameras located on the vehicle, providing a 360$^\circ$ field of view. All cameras in each scene have extrinsic $(R,t)$ and intrinsic $K$ calibration parameters provided at every timestamp; the intrinsic parameters remain unchanged with time. Other perception sensors in the nuScenes dataset (radar and lidar) are not used in this work.

\subsection{Image Encoder}
We use the simple and effective encoder-decoder architecture for map-view semantic segmentation from \cite{zhou2022cross}. In summary, the authors proposed an image encoder that generates a multi-scale feature representation $\{\phi\}$ for each input image, which is then combined into a shared map-view representation using a cross-view cross-attention mechanism. This attention mechanism utilizes a positional embedding $\{\delta\}$ to capture both the geometric structure of the scene, allowing for accurate spatial alignment, and the sequential information between different camera views, facilitating temporal understanding and context integration. All camera-aware positional embeddings are presented as a single key vector $\delta = [\delta_1, \delta_2......\delta_6]$. Image features are combined into a value vector $\phi = [\phi_1, \phi_2.....]$. Both are merged to create a comparison of attention keys and subsequently, a softmax-cross attention is used \cite{vaswani2017attention}.

\subsection{Cross Attention}
As illustrated in Figure \ref{fig:image2}, the cross-view transformation component aims to establish a connection between a map view and image features, as presented by \cite{zhou2022cross}. To summarise, precise depth estimation is not learned; rather, the transformer learns a depth proxy through positional embedding $\{ \delta\}$ ($x^{world}$  remains ambiguous). The cosine similarity is used to express the geometric relationship between the world and unprojected image coordinates:

\begin{equation} \label{eqn:cossim}
\cos(\theta) = \frac{\left(R_k^{-1}K_k^{-1}x^{image}\right) \cdot \left(x^{world} - t_k\right)}{\lVert R_k^{-1}K_k^{-1}x^{image} \rVert \lVert x^{world} - t_k \rVert}
\end{equation}

where denoted as $x^{image} \in \mathbb{P}^{3}$ is a homogeneous image point for a given world coordinate $x^{world} \in \mathbb{R}^{3}$. The cosine similarity traditionally relies on precise world coordinates.

However, in this approach, the cosine similarity is augmented with positional embeddings, thus having the capability to learn both geometric and appearance features \cite{zhou2022cross}. Direction vectors $d_{k,i} = R_k^{-1} K_k^{-1} x^{image}_i$ are created for each image coordinate $x^{image}_i$, serving as a reference point in world coordinates. 
An MLP is used to convert the direction vector $d_{k,i}$ into a $D$-dimensional positional embedding denoted as $\delta_{k,i} \in \mathbb{R}^D$ (Per \cite{zhou2022cross}, we have set the value of $D$ to 128).

\subsection{Joint Vehicle Segmentation}
To enhance the vehicle segmentation, we have designed our segmentation head to be simple, utilizing a series of convolutions on the bird's-eye view (BEV) feature. Specifically, it consists of four $3\times3$ convolutions followed by a $1\times1$ convolution, resulting in a BEV tensor of size $h\times w\times n$, where $n$ represents the number of categories. In our case, we set $n$ to 1, as we focus solely on the vehicles and other agents related to it following the approach used in the cross-view transformer \cite{zhou2022cross}. To enhance road and vehicle segmentation in the dataset using an encoder-decoder transformer, we employ the following equation:
\begin{equation*}
 y = f(X1, X2) 
\end{equation*}
where y is the output segmentation map, $X1$ is the input image from one sensor modality (e.g., camera), and $X2$ is the input image from another sensor modality (e.g., map information). $f$ is the cross-view transformer, which learns to combine the information from the two modalities to produce a more accurate segmentation map. The cross-attention mechanism can be implemented using the following equation:
\begin{equation}
M = \text{softmax}\left(\frac{Q.(K^T)}{\sqrt{d_k}}\right)V
\end{equation}
where $Q$, $K$, and $V$ are the queries, keys, and values, respectively, for each modality. The dot product between the queries and keys is present in the form of ${Q.(K^T)}$ \SKS{The dot product between queries and keys } \AD{Where is the dot product in the above equation?} is divided by the square root of the dimensionality of the key vectors $(d_k)$ to prevent the dot product from becoming too large. Subsequently, the obtained attention weights are employed to weigh the values associated with each modality. These weighted values are then combined to generate the output feature map $M$.

\subsection{Spatio-Temporal Probabilistic Network (STPN)}

This section describes the Spatio-temporal probabilistic network for trajectory prediction of the future states of an ego vehicle and a high-definition map, assuming that we have access to the state outputs of an object detection and tracking system of sufficient quality for autonomous vehicles, based on \cite{phan2021contract}. The agents that an ego vehicle interacts with at time $t$ are denoted by the set $I_t$, and $s^i_t$ represents the state of agent $ i \in I_t$ at time $t$. The discrete-time trajectory of agent $i$ for times $t = \big(m,.....,n)$ is denoted by $s^i_{m:n}$ = $\big[s^i_m,......,s^i_n]$, where $m < n$ and $i \in I_t$. \\

Additionally, we presume that the high-definition map, as depicted in our proposed method, will be accessible. This includes lane geometry, crosswalks, drivable areas, and other pertinent information.
The scene context over the past $m$ steps, which includes the map and partial history of ego vehicles, is denoted by ${C = \bigl\{\bigcup_i s^i_{t-m:t}; \texttt{Map Information}\bigl\}}$.

Our architecture follows the trajectory prediction layer with the approach presented in \cite{cui2019multimodal}. To achieve effectiveness in this domain, we employ ResNet-50 (Table:\ref{table:1}) \cite{7780459}, as recommended by previous research \cite{cui2019multimodal,chai2019multipath}. Although our network currently generates predictions for one agent at a time, our approach has the potential to predict for multiple agents simultaneously in a manner similar to \cite{chai2019multipath}. However, we limit our focus to single-agent predictions (as in \cite{cui2019multimodal}) to streamline the paper and emphasize our primary contributions. To represent probabilistic trajectory predictions in multiple modes, we utilize a classification technique that selects the relevant trajectory set based on the agent of interest and scene context $C$. The softmax distribution is employed, as is typical in classification literature. Specifically, the probability of the $k$-th trajectory is expressed as follows:

\begin{equation}
 p(s^k_{t:t+N}| x) = \frac{\texttt{exp} f_k(x)}{\sum_{i}\texttt{exp} f_i(x)} 
\end{equation}
where $f_i(x) \in \mathbb{R}$ is the output of the network of probabilistic  layer.
We have implemented Multi-Trajectory Prediction (MTP) \cite{cui2019multimodal} with adjustments made for our datasets. This model forecasts a set number of trajectories (modes) and determines their respective probabilities. Note that we are now focusing on single trajectory prediction (STP)\cite{djuric2020uncertainty}.

\subsection{Loss Function}
The loss function employed for vehicle segmentation in our transformer-based model is defined as follows:

\begin{equation}
\begin{split}
\mathcal{L}_{\text{seg}}(\mathbf{m}, \mathbf{\hat{m}}) = -\frac{1}{N} \sum_{i=1}^{N}  \bigl[ &m_i \cdot \log(p(\hat{m}_i)) \\
&+ (1 - m_i) \cdot \log(1 - p(\hat{m}_i)) \bigr]
\end{split}
\end{equation}

where, $\mathcal{L}_{\text{seg}}(\mathbf{m}, \mathbf{\hat{m}})$ is the binary cross-entropy loss \cite{jadon2020survey} for vehicle segmentation, $\mathbf{m}$ is the input tensor, and $\mathbf{\hat{m}}$ is the target tensor for all $N$ points. This loss function is particularly valuable for binary classification challenges where our model generates logits (unbounded real numbers) as output. It facilitates the computation of the binary cross-entropy loss concerning binary target labels $\mathbf{\hat{m}}$, ensuring effective training and performance evaluation for vehicle segmentation in our transformer-based approach.

In terms of trajectory prediction, the loss function we are considering is one of the most commonly used: the mean squared error (MSE). This loss function typically involves measuring the dissimilarity between the predicted and the ground-truth trajectories.
\vspace{-1mm}
\begin{equation}
  \mathcal{L}\mkern 2mu_{traj} = \frac{1}{N}\sum_{i=1}^{N}||\hat{y}_i - y_i||^2_2
\end{equation}

Here, $N$ is the number of training examples, $\hat{y}_i$ is the predicted trajectory for ego vehicle $i$, and $y_i$ is the corresponding ground truth trajectory. The squared difference between the two trajectories is calculated element-wise and then averaged across all elements in the trajectory. The resulting value is the mean squared error loss, which measures the overall performance of the model in predicting the trajectories for the ego vehicle.

Our final loss function $\mathcal{L}_{\text{total}}$ constitutes two components, as shown in the equation below.

\begin{equation}
\begin{aligned}
\mathcal{L}_{\text{total}} = \alpha \mathcal{L}_{\text{seg}} + \beta \mathcal{L}_{\text{traj}}
\end{aligned}
\end{equation}
Gradients are mutually shared by both tasks till the initial layers of the network. In the above equation, $\alpha$ and $\beta$ are the hyperparameters to balance between segmentation and trajectory prediction losses.
\vspace{-1mm}

\section{EXPERIMENTAL SETUP}
\label{sec:results}
\subsection{Dataset}
Experiments are carried out on the nuScenes dataset \cite{caesar2020nuscenes}, which comprises $1000$ video sequences gathered in Boston and Singapore. The dataset is composed of scenes that have a duration of 20 seconds and consist of 40 frames each, resulting in a total of 40k samples. The dataset is divided into training, validation, and testing sets, with 700, 150, and 150 scenes respectively. The recorded data provides a comprehensive $360^{\circ}$ view of the surrounding area around the ego-vehicles and comprises six camera perspectives. Note that we are employing identical train-test-validation splits as those used in the previous works \cite{zhou2022cross,philion2020lift} for comparison.

\subsection{Transformer Architecture and Implementation Details}

The initial step of the network involves creating a camera-view representation for each input image. To achieve this, we utilize EfficientNet-B4 \cite{tan2019efficientnet} as the feature extractor and input each image $I_i$ to obtain a multi-resolution patch embedding  ${\bigl\{\delta^1_1, \delta^2_1,\delta^3_1,.....\delta^R_n\bigl\}}$, where $R$ denotes the number of resolutions that are taken into account.

According to our experimental findings, accurate results can be achieved when using $R=1$ resolution. However, if we were to increase the value of $R$ to $2$, as suggested by CVT in \cite{zhou2022cross}, the camera-view representation for each input image in the network would incorporate additional information, such as BEV features. While this has the potential to result in a more detailed representation of the input images, it also comes with drawbacks, including increased computational requirements and a higher risk of overfitting. \\

The processing for each resolution is carried out individually, beginning with the lowest resolution. We employ cross-view attention to map all image features to a map-view and refine the map-view embedding, repeating this procedure for higher resolutions. In the end, we employ three up-convolutional layers to produce the output at full resolution. Once we obtain the full-resolution output, we input the ego vehicle features, which have a resolution of $h_{bev}\times w_{bev}\times 256$, into the probabilistic function for trajectory forecasting, resulting in the set of trajectories $[p_1,p_2,p_3,...,p_n]$. Subsequently, we refine and obtain the probabilistic value, which represents our final trajectory.

To implement the architecture, we employ a pre-trained EfficientNet-B4 \cite{tan2019efficientnet} that we fine-tune. The two scales, $(28, 60)$ and $(14, 30)$, correspond to an $8\times$ and $16\times$ downscaling, respectively. For the initial map view positional embedding, we use a tensor of learned parameters with dimensions $ w \times h \times D$, where $D$ is set to $128$. To ensure computational efficiency, we limit the grid size to $ w = h = 25$, as the cross-attention function becomes quadratic in growth with increasing grid size. The encoder comprises two cross-attention blocks, one for each scale of patch features, which utilize multi-head attention with 4 heads and an embedding size of $d_{head} = 64$. \\

The decoder includes three layers of bilinear upsampling and convolution, each of which increases the resolution by a factor of 2 up to the final output resolution of $200 \times 200$, corresponding to a $100 \times 100$ meter area around the ego-vehicle. The map-view representation obtained through the cross-attention transformer is passed through the joint vehicle segmentation module to accurately identify the vehicle's segmentation. This segmentation is then utilized as input to the Spatial-Temporal Probabilistic Network (STPN), which offers probabilistic predictions. Instead of providing a single deterministic trajectory, the network offers a probability distribution over possible future trajectories. This information aids in identifying the motion planning of the ego vehicle. Precisely segmenting the pixels corresponding to the ego vehicle enables the system to more accurately estimate its position, speed, and orientation in relation to other objects in the environment. This, in turn, facilitates improved decision-making during navigation. Figure \ref{fig:image2} offers a comprehensive overview of this architecture.

\vspace{-5mm}

\section{ABLATION STUDY}

We perform a detailed ablation experiment to assess the influence of several factors on the functionality of our segmentation model. We specifically examined the impacts of various backbone models and loss functions.


\begin{table}[ht]
\caption{Comparison study of \textbf{different standard backbone models employed for trajectory prediction} on nuScenes dataset \cite{caesar2020nuscenes}}\label{table:1}
\centering
\scalebox{0.9}{
\begin{tabular}{c|ccc}
  \hline
  \textbf{Backbone} &  \textbf{\# Params. (M)} & \textbf{Features} & \textbf{MSE $\downarrow$} \\
  \hline
  \hline
  \textit{EfficientNet-80}  & 1.9 & 1280 & 0.3385 \\
  \textit{DenseNet-121}  & 1.7 & 1024 & 0.2079 \\ 
  \textit{ResNet-50}  & \textbf{1.4} & \textbf{512} & \textbf{0.1062}  \\
  \hline
\end{tabular}
}
\end{table}

We performed an ablation on different backbone models to investigate their impact on the performance of our target task on the nuScenes dataset, as presented in Table \ref{table:1}. Notably, the ResNet-50 backbone, with 1.4 million trainable parameters and a feature size of 512, demonstrated promising results, achieving the lowest MSE of 0.1062. It is likely that ResNet-50 works well for trajectory prediction on the nuScenes dataset, as its model parameters align well with the characteristics of that dataset.

\begin{table}[ht]
\caption{Ablation on \textbf{different loss functions for segmentation task} on the nuScenes dataset \cite{caesar2020nuscenes} }\label{table:2} \centering
\begin{tabular}{c|cc}
  \hline

\textbf{Loss Function} &  \textbf{No. of Class} & \textbf{Loss} $\downarrow$ \\
  \hline
  \hline
  \textit{Binary Cross Entropy}  & 2 & 0.1848  \\
  \textit{Binary Focal Loss}  & 2 &  0.2758 \\ 

  \hline
\end{tabular}
\end{table}

In our task, we utilize the binary cross-entropy loss function, which aligns well with the inherent characteristics of our standard binary classification problem. Additionally, we explore and compare alternative loss functions, including binary focal loss. However, our findings indicate that the binary cross-entropy loss function yields superior results, as presented in Table \ref{table:2}. This is primarily attributed to the balanced distribution of classes within our dataset, which favors the effectiveness of binary cross-entropy in accurately modeling the classification problem.


\begin{table*}[ht]
\caption{Comparison of \textbf{visibility-based methods for Setting 1 and Setting 2}, where our method achieves the highest visibility rate among those with visibility greater than $40\%$.}\label{table:3} \centering
\begin{tabular}{ccc}
  \hline
 \multicolumn{1}{c}{} & \multicolumn{2}{c}{Visibility $>$ $40\%$} \\

\textbf{Method} & Setting 1   & Setting 2\\
  \hline
    \textit{LSS \cite{philion2020lift}} & - & 32.1   \\ 
    \textit{CVT \cite{zhou2022cross}} & 37.5 & 36.0 \\ \hline
  \textit{\textbf{BEVSeg2TP (Ours)}} & \textbf{37.8} & \textbf{37.9} \\ 
  \hline
\end{tabular}
\end{table*}

\begin{table*}[ht]
\caption{Comparison of \textbf{vehicle segmentation performance on the nuScene dataset} using different methods, including LSS, CVT, and our proposed method. Results are presented in terms of Intersection over Union (IoU) scores.}\label{table:4}
\centering
\begin{tabular}{c|cc}
  \hline
  
  \textbf{Method} &  Resolution \textbf{R} & Vehicle $\uparrow$\\ 
  \hline
  \hline
  \textit{LSS \cite{philion2020lift}} & -  & 32.1\\ 
  \textit{CVT \cite{zhou2022cross}} & 2   & 36.0 \\ \hline
 \textit{\textbf{BEVSeg2TP (Ours)}} & 1  & \textbf{37.9} \\ 
  \hline
\end{tabular}
\end{table*}

\begin{table*}[h!]
\caption{Comparison of the \textbf{Minimum Average Prediction Error (MinADE) and Final Displacement Error (MinFDE) }for Competing Methods on the nuScenes Dataset, over a Prediction Horizon of 6 Seconds.}\label{table:5} \centering
\scalebox{0.83}{
\begin{tabular}{c|cccccc}
  \hline

\textbf{Method} &  MinADE$_{5}$ $\downarrow$ & MinADE$_{10} \downarrow$  & MinADE$_{15} \downarrow$  & MinFDE$_{5}$ $\downarrow$ & MinFDE$_{10}$ $\downarrow$ & MinFDE$_{15}$ $\downarrow$\\
  \hline
  \hline
\textit{Const Vel and Yaw} & 4.61 & 4.61 & 4.61 & 11.21 & 11.21 & 11.21 \\
  
\textit{Physics oracle} & 3.69 & 3.69 & 3.69 & 9.06 & 9.06 & 9.06\\

\textit{CoverNet \cite{phan2020covernet}} & 2.62 & 1.92 & 1.63 & 11.36 & - & - \\
\textit{Trajectron++ \cite{salzmann2020trajectron++}} & 1.88 & 1.51 & - & - & - & - \\ 
\textit{MTP \cite{cui2019multimodal}} & 2.22 & 1.74 & 1.55 & 4.83 & 3.54 & 3.05\\ 
\textit{MultiPath \cite{chai2019multipath}} & 1.78  & 1.55 & 1.52 & \textbf{3.62} & 2.93 & 2.89 \\
\hline

\textit{\textbf{BEVSeg2TP (Ours)} }  & \textbf{1.63} & \textbf{1.29} & \textbf{1.15} & 3.85 & \textbf{2.13} & \textbf{1.65} \\ \hline
\end{tabular}
}
\end{table*}
\vspace{-5mm}

\section{RESULTS}

We evaluate the BEV map representation and trajectory planning of the BEVSeg2TP model on the publicly available nuScenes dataset. 
The evaluation is conducted in two different settings - 'Setting 1' refers to a $100m \times 50m$ grid with a $25cm$ resolution, while 'Setting 2' refers to a $100m \times 100m$ grid with a $50cm$ resolution. During training and validation, vehicles with a visibility level above the predefined threshold of $40\%$ are considered. Table \ref{table:3} demonstrates the comparison of our proposed approach with other existing works such as LSS \cite{philion2020lift} and CVT \cite{zhou2022cross}.

First, we compare the BEV segmentation obtained from various methods, including LSS and CVT with the results from our proposed BEVSeg2TP. Accurately predicting the future motion of vehicles is critical, as it helps the model gain a comprehensive understanding of the environment by capturing the spatial relationships among pedestrians, vehicles, and obstacles. However, our second contribution focuses on improving map-view segmentation of vehicles. Our experimental findings show that employing a resolution of $R=1$ yields promising results. However, increasing the value of $R$ to $2$, as recommended by CVT, would lead to the camera-view representation for each input image in the network losing information, such as BEV features.  We conducted further evaluations using various methods, as illustrated in Table \ref{table:4}.

\begin{figure*}[ht]
    \centering
    \includegraphics[width=0.95\textwidth]{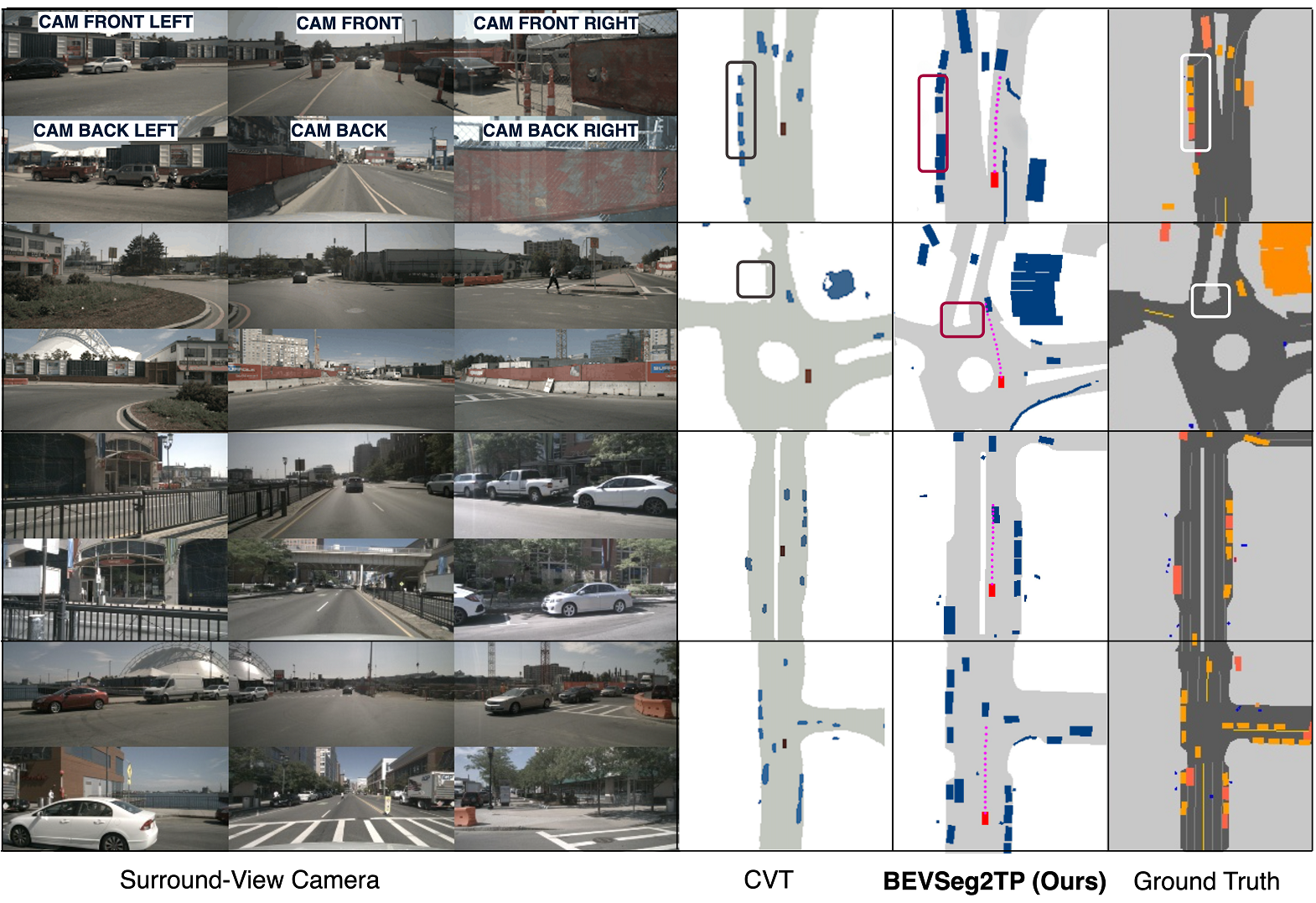}
    \caption{\textbf{Qualitative results of BEVSeg2TP model for joint vehicle segmentation and ego vehicle trajectory prediction:} Six camera views around the vehicle (top three facing forward, bottom three facing backwards) with ground truth segmentation on the right. Our trajectory prediction with improved map-view segmentation (second from right) compared to the CVT  method (third from right).}
    \label{fig:image3}
\end{figure*}

As shown in Table \ref{table:5}, the ablation study has been evaluated by comparing it with four baselines:  \cite{cui2019multimodal} \cite{chai2019multipath} \cite{phan2020covernet} and \cite{salzmann2020trajectron++} and two physics-based approaches. These four baselines are a recently proposed model which is considered to be the current state-of-the-art for multimodel trajectory prediction. This comparison aims to assess the effectiveness and accuracy of our model in predicting trajectories in comparison to existing models. The goal is to determine if our model performs better than or at least as well as the state-of-the-art baseline model. By doing so, we can gain insight into the strengths and weaknesses of our model and identify areas for further improvement. To evaluate the performance of our model on the nuscenes dataset, we first obtained the output trajectories $\big[{y_1,y_2,y_3,....y_n}\big]$. We evaluated the performance of the model on this specific dataset for different values of $K$, where $K$ was set to 5, 10, and 15 respectively.
\begin{equation}
\mathbf{MinADE_k} =  \min_{i\in\{1...K\}} \frac{1}{T_f} \sum_{t=1}^{T_f} \left\Vert y_t^{\text{gt}} - y_t^{(i)} \right\Vert_2
\end{equation}

To train the model, we minimized the minimum average displacement error over $K$ (MinADE${_k}$) on the training set. In other words, we aimed to reduce the error between the predicted trajectories and the actual trajectories by minimizing the minimum distance between them for each of the $K$ time steps. This method allowed us to improve the accuracy of our model's predictions and ensure that it performs well on the nuScenes dataset. Here, $y_{t}^{gt}$ represents the ground truth position of the object at the final time step T, and $y_{t}^{(i)}$ represents the predicted position of the object at the final time step T for the $i_{th}$ trajectory in the set of $K$ trajectories. \\

We took the output trajectories $\big[{y_1,y_2,y_3,....y_n}\big]$ and we used $K=15 $ for nuscenes datasets. we minimize the minimum over $K$ average displacement error (MinADE${_k}$) over the training set.
As depicted in Figure \ref{fig:image3}, on the left-hand side of the image, there are six camera views surrounding the vehicle. The top three views are oriented forward, while the bottom three views face backwards. On the right side of the image, there is ground truth segmentation for reference. Moving from right to left, the second image from the right displays our trajectory prediction, along with improved map-view segmentation for vehicles. Lastly, the third image from the right illustrates the CVT \cite{zhou2022cross} method, which we use to conduct a comparison and present the results. \\
The \textbf{black} color corresponds to the results obtained using a model called CVT, the \textcolor{red}{red} color corresponds to the results obtained using our model, and the \textcolor{gray}{white} color corresponds to the nuScenes ground truth, which is the true segmentation of the images. The purpose of the comparison was to evaluate the performance of the other model and compare it with our model. Figure \ref{fig:image3} reveals that our model performs well compared to the other model in both vehicle and road segmentation tasks. When it comes to vehicle segmentation, our model demonstrates a high level of accuracy in identifying the precise positions of vehicles within the image. In contrast, the other model exhibits a slightly lower level of accuracy in this regard. 
This distinction is clearly visible in the accompanying figure, where the red markings, representing the outcomes produced by our model, closely align with the green markings, representing the ground truth, in comparison to the black markings, which correspond to the results generated by the other model. Similarly, with regard to road segmentation, our model also exhibits decent performance. To gain further insights, additional results can be explored via the following link: \textcolor{purple}{\href{https://youtu.be/FNBMEUbM3r8}{https://youtu.be/FNBMEUbM3r8}}.

\section{CONCLUSION} 
In this paper, we propose BEVSeg2TP - a surround-view camera bird's-eye-view-based joint vehicle segmentation and ego vehicle trajectory prediction using encoder-decoder transformer-based techniques that have shown promising results in achieving safe and effective driving for autonomous vehicles. The system processes images from multiple cameras mounted on the vehicle, performs semantic segmentation of objects in the scene, and predicts the future ego vehicle trajectory of surrounding vehicles using a combination of transformer and spatio-temporal probabilistic network (STPN) to calculate the trajectory. The predicted trajectories are projected back to the ego vehicle's bird's-eye-view perspective, providing a comprehensive understanding of the surrounding traffic dynamics. Our findings underscore the potential benefits of employing transformer-based models in conjunction with spatio-temporal networks, highlighting their capacity to significantly enhance trajectory prediction accuracy. Ultimately, these advancements contribute to the overarching goal of achieving a safer and more efficient autonomous driving experience.
\\
While the camera configuration of nuScenes is important, it is not a typical commercially deployed surround-view system. Commercial surround view systems, used for both viewing and vehicle automation and perception tasks \cite{Kumar2023,Eising2022}, typically employ a set of four fisheye cameras around the vehicle. In the future, we intend to apply the methods discussed here to fisheye surround-view camera systems.

\section*{\uppercase{Acknowledgements}}

This publication has emanated from research supported in part by a grant from Science Foundation Ireland under Grant number 18/CRT/6049. For the purpose of Open Access, the author has applied a CC BY public copyright licence to any Author Accepted Manuscript version arising from this submission.

\bibliographystyle{apalike}
{\small
\bibliography{example}}



\end{document}